\documentclass{article} 
\usepackage{iclr2027_conference,times}

\usepackage{amsmath,amsfonts,bm}

\def\eqref#1{equation~\ref{#1}}

\def\1{\bm{1}}

\DeclareMathAlphabet{\mathsfit}{\encodingdefault}{\sfdefault}{m}{sl}
\SetMathAlphabet{\mathsfit}{bold}{\encodingdefault}{\sfdefault}{bx}{n}

\usepackage{hyperref}
\usepackage{url}

\usepackage{hyperref}
\usepackage{url}
\usepackage{graphicx}
\usepackage{booktabs}
\usepackage{multirow}
\usepackage{graphicx}

\usepackage{wrapfig}
\usepackage{booktabs}
\usepackage{multirow}
\usepackage{graphicx}
\usepackage{booktabs}
\usepackage{makecell}
\usepackage{xcolor}
\usepackage{pifont}
\usepackage{ragged2e}

\usepackage{listings}
\usepackage{xcolor}

\usepackage{arydshln}

\lstdefinestyle{promptstyle}{
    basicstyle=\ttfamily\footnotesize,
    breaklines=true,
    breakatwhitespace=false,
    columns=fullflexible,
    frame=single,
    framesep=5pt,
    rulecolor=\color{black!50},
    backgroundcolor=\color{black!2},
    showstringspaces=false,
    keepspaces=true,
    tabsize=2,
    aboveskip=8pt,
    belowskip=8pt
}

\usepackage{listings}
\lstdefinestyle{jsonexample}{
  basicstyle=\ttfamily\scriptsize,
  breaklines=true,
  breakatwhitespace=true,
  columns=fullflexible,
  keepspaces=true,
  showstringspaces=false,
  frame=single,
  framerule=0.4pt,
  rulecolor=\color{black!35},
  backgroundcolor=\color{black!2},
  xleftmargin=0.5em,
  xrightmargin=0.5em
}

\title{ModularRSI: Modular and Generalizable Recursive Harness Self-Improvement}

\author{
    ~ \\[-3ex]
  \begin{minipage}{0.95\textwidth}
    \raggedright
    {\bf
    Siwei Wu\textsuperscript{2,3,*},
    Jincheng Ren\textsuperscript{4,6,*},
    Yizhi Li\textsuperscript{3,$\dagger$},
    Haau-Sing Li\textsuperscript{3},
    Chengran Yang\textsuperscript{3},
    Xudong Ma\textsuperscript{3},
    Yuxuan Zhang\textsuperscript{3},
    Weicheng Gu\textsuperscript{3},
    Jian Yang\textsuperscript{1,$\ddagger$},
    Riza Batista-Navarro\textsuperscript{2},
    Chuanyi Zhang\textsuperscript{6},
    Xianglong Liu\textsuperscript{1},
    Ming Zhou\textsuperscript{5},
    Bryan Dai\textsuperscript{3},
    Chenghua Lin\textsuperscript{2$\ddagger$}
    }
  \end{minipage}
  \\[0.7cm]
  \begin{minipage}{0.95\textwidth}
    \raggedright
    \textsuperscript{1}Beihang University,
    \textsuperscript{2}University of Manchester,
    \textsuperscript{3}IQuest Research, \\
    \textsuperscript{4}M-A-P,
    \textsuperscript{5}Langboat,
    \textsuperscript{6}Hohai University\\[0.12cm]
    \texttt{\{siwei.wu-2@postgrad., chenghua.lin@\}manchester.ac.uk}
  \end{minipage}
}

\iclrfinalcopy
\begin{document}

\maketitle

\begingroup
\renewcommand{\thefootnote}{\fnsymbol{footnote}}
\footnotetext[1]{Equal contribution.}
\footnotetext[2]{Project leader.}
\footnotetext[3]{Corresponding authors.}
\endgroup

\begin{abstract}

Recent work extends Recursive Self-Improvement (RSI) to agent harnesses for long-horizon coding and terminal tasks, enabling agents to improve their execution mechanisms from experience. However, achieving \textbf{generalizable harness RSI} faces two challenges: evolution on downstream benchmarks makes transferable improvements difficult to distinguish from benchmark-specific adaptation, while coarse task-level outcomes provide limited guidance on \textbf{where that change should be implemented} and \textbf{what execution behavior should change} in the harness.
We propose \textbf{ModularRSI}, a modular framework for generalizable harness self-evolution that addresses both \textbf{mechanism-level localization} and \textbf{behavioral-level attribution} in credit assignment.
ModularRSI decomposes the harness into five functional modules, including \textit{Agent Loop}, \textit{Tool Use}, \textit{Observation Management}, \textit{Context Management}, and \textit{Task Completion Detection}, and evolves each module independently to enable localized mechanism improvement.
Within each module, ModularRSI performs contrastive analysis between successful and failed trajectories from the same task and aggregates evidence across tasks to identify recurring execution deficiencies.
Finally, the independently evolved modules are integrated to construct the final harness.
We further curate 2,000 executable evolution tasks from external sources that are disjoint from downstream evaluation benchmarks.
Experiments on TerminalBench (TB) 2.0 and SWE-Bench-Verified demonstrate consistent generalization across unseen tasks, evolution domains, and foundation models. Using DeepSeek-V4-Flash-Preview, in-domain evolution improves accuracy from 47.57\% to 52.43\% on TB 2.0 and from 73.40\% to 76.40\% on SWE-Bench-Verified. Beyond in-domain settings, the evolved harness also transfers effectively across evolution domains and foundation models, yielding approximately 2\%--5\% performance gains. Under a unified benchmark-disjoint protocol and matched evolution budgets, ModularRSI achieves 52.43\% accuracy on TerminalBench 2.0, outperforming Meta-Harness at 49.44\% and AHE at 49.81\%. Ablation studies further confirm that both modular evolution and contrastive analysis contribute to these improvements.
All code and datasets are available at \url{https://github.com/IQuestLab/ModularRSI}.

\end{abstract}

\section{Introduction}

CLI agents have achieved remarkable performance on complex software engineering and terminal-based tasks~\citep{jimenez2024swe,deng2025swe,merrill2026terminalbenchbenchmarkingagentshard,hong-etal-2026-hiras}. Beyond foundation models, their effectiveness increasingly depends on agent harnesses that govern execution, tool interaction, context management, and environment feedback. Recent work has therefore explored recursive self-improvement (RSI) of harnesses, allowing agents to refine these execution mechanisms from experience.

However, the central challenge of \textbf{generalizable harness RSI} is not whether a harness can improve on the experience used for evolution, but whether the resulting changes represent \textbf{transferable improvements to the underlying execution mechanism}. Existing methods  (e.g., Meta-Harness, AHE, and HarnessFix)  often evolve directly on downstream benchmarks or subsets derived from them~\citep{lee2026meta,lin2026agentic,du2026livingharnessinteractiveagentevolver,chen2026failed,pan2026evolvingagentsdarkretrospective,nguyen2026recursiveselfevolvingagentsheldout}, making reusable improvements difficult to distinguish from benchmark-specific adaptation. Even under separated evolution and evaluation, task-level rewards provide only coarse supervision: individual trajectories entangle systematic harness deficiencies with task-specific reasoning, while broad whole-harness updates make it difficult to localize reusable improvements without affecting unrelated mechanisms.
We therefore view \textbf{generalizable harness RSI as a two-level credit-assignment problem}, which becomes particularly important under benchmark-disjoint evolution.
A successful self-improvement procedure must answer two complementary questions: \textbf{where in the harness should that improvement be implemented} and \textbf{what execution behavior should be improved}. 
The first is a \textbf{mechanism credit-assignment problem}: once a reusable deficiency has been identified, the corresponding change must be localized to the responsible harness components rather than broadly modifying unrelated components.
The second is a \textbf{behavioral credit-assignment problem}: execution trajectories must be compared and aggregated to distinguish recurring harness deficiencies from task-specific decisions or stochastic failures. 
Our central hypothesis is that transferable harness self-improvement requires improving the precision of credit assignment at both levels, rather than simply enlarging the self-modification space.
HarnessFix~\citep{chen2026failed} improves failure localization from execution trajectories, but stops short of explicitly assigning the identified deficiencies to the responsible harness mechanisms. AHE~\citep{lin2026agentic} recognizes functionally distinct components within the harness, yet still evolves them within a joint harness-level process rather than independently attributing and optimizing mechanism-specific improvements. Consequently, behavioral and mechanism-level credit assignment remain only partially addressed.

To address these two levels of credit assignment, we propose ModularRSI, a modular approach for generalizable harness evolution. For mechanism-level credit assignment, it decomposes the evolvable harness into five functionally coherent modules (i.e., Agent Loop, Tool Use, Observation Management, Context Management, and Task Completion Detection). Each module is evolved independently under a restricted modification scope, followed by cross-module integration into a unified harness. For behavioral-level credit assignment, it contrasts successful and failed trajectories within the same task and aggregates evidence across tasks to identify recurring deficiencies beyond task-specific variation, while guiding localized updates to the responsible harness mechanisms.

To enable a clean evaluation of generalizable harness RSI, we establish a \textbf{benchmark-disjoint evolution protocol} and curate 2,000 executable tasks from external sources that are disjoint from downstream benchmarks. Experiments on \textbf{TerminalBench 2.0} and \textbf{SWE-Bench Verified} show consistent generalization across unseen tasks, domains, and foundation models, while controlled studies demonstrate the benefits of modular evolution, contrastive trajectory analysis, and some other components.

Our main contributions are summarized as follows:
\begin{enumerate}
    \item We formulate \textbf{generalizable harness RSI as a benchmark-disjoint credit-assignment problem} and propose \textbf{ModularRSI}, which performs credit assignment at two complementary levels. Contrastive trajectory analysis identifies recurring behavioral deficiencies, while module-restricted evolution localizes and implements the corresponding mechanism-level improvements.
    
    \item We establish a \textbf{benchmark-disjoint evolution protocol} and independently curate 2,000 executable evolution tasks from external data sources. Through executable validation, manual quality control, and similarity screening, this resource enables harness self-improvement to be studied without direct exposure to downstream benchmark instances.
    
    \item We extensively evaluate ModularRSI on \textbf{TerminalBench 2.0} and \textbf{SWE-Bench Verified}. The evolved harness generalizes across unseen tasks, evolution domains, and foundation models. Controlled studies further show that modular evolution substantially outperforms joint and non-modular evolution under matched budgets, while removing contrastive attribution degrades performance, supporting the importance of both behavioral and mechanism-level credit assignment.
\end{enumerate}

\section{Related Work}

\subsection{Agent Harnesses and Self-Improvement}

Alongside advances in coding models~\citep{yang2026iquestcoderv1technicalreport,yang-etal-2026-loopcoder}, executable environments and verified trajectories provide scalable signals for agent improvement~\citep{wu2026large,gandhi2026endless,cheng2026terminal}.
Recent work uses trajectory feedback to optimize prompts and external agent mechanisms~\citep{agrawal2026gepa,zhang2026rewardharness,fan2026medclaw}, diagnose failures and localize editable components~\citep{lin2026agentic,chen2026failed}, or directly evolve agent by synthesizing harness code~\citep{zhang2026darwin,lou2026autoharness,lee2026meta,ren2026selfevolvingframeworkefficientterminal}.

\subsection{Generalization in Self-Improvement}

A central challenge is distinguishing transferable improvements from adaptation to development data.
HarnessDev and related studies examine harness evolution, feedback budgets, and held-out generalization~\citep{wu2026harnessdev,wang2026rethinking}.
Aspire studies weight and harness evolution under hidden evaluation~\citep{wu2026aspire}, while S$^3$Gym separates self-testing, self-judging, and improvement through history, memory, or parameter updates~\citep{shi2026s3gym}.
Despite these advances, existing harness-evolution methods often rely on benchmark-derived data or treat the harness as a monolithic whole, leaving generalizable, fine-grained improvement underexplored.
We address this gap with a benchmark-disjoint evolution dataset and protocol, and propose ModularRSI, which uses contrastive trajectory analysis to guide independent, module-restricted evolution followed by cross-module integration.

\begin{figure*}[t]
    \centering
    \includegraphics[width=0.75\textwidth]{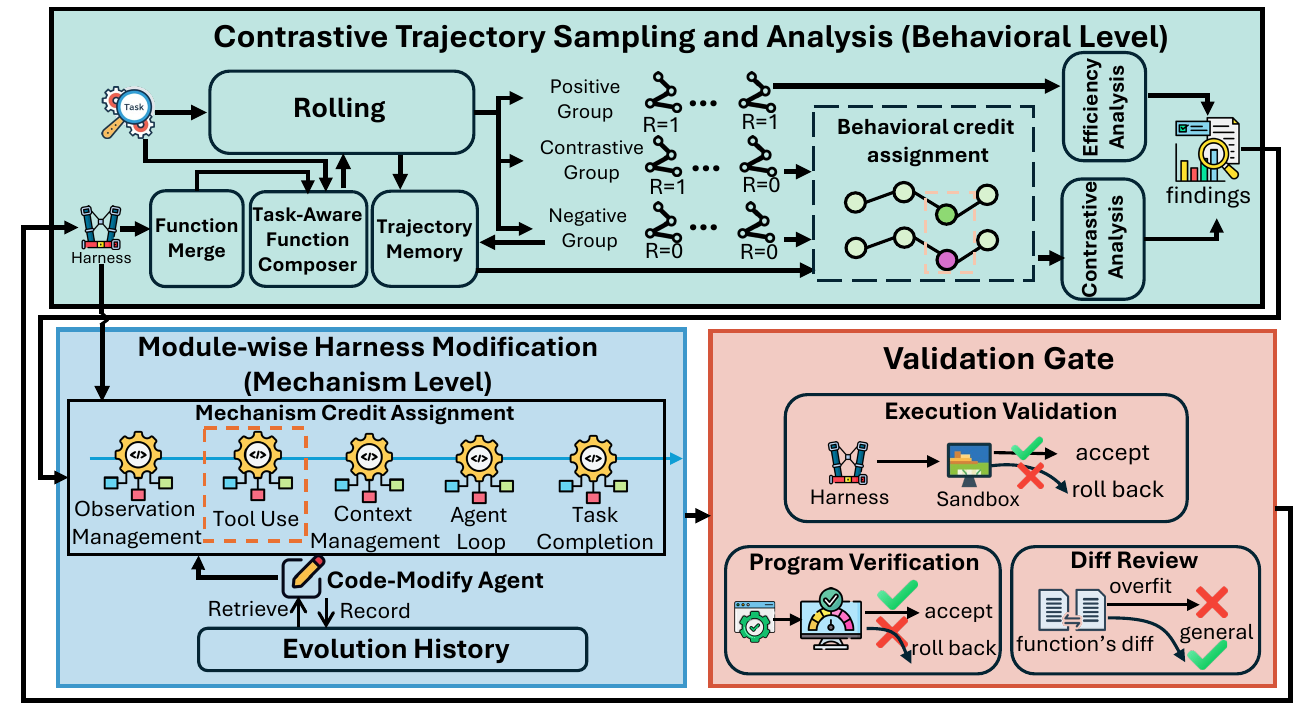}
    \caption{
        Overview of ModularRSI. The framework collects and analyzes
        contrastive trajectories, performs module-wise harness modification, and
        retains modifications through validation gates.
    }
    \label{fig:main_method}
    \vspace{-2mm}
\end{figure*}

\section{Method}
\label{Sec:Method}

\subsection{Overview of ModularRSI}

As illustrated in Fig.~\ref{fig:main_method}, ModularRSI organizes harness self-improvement through \textbf{module-restricted evolution} and subsequent \textbf{cross-module integration}.
It independently evolves five functional modules from the same initial harness without sharing intermediate updates.
Within each module-specific lineage, contrastive trajectory attribution identifies behavioral deficiencies and guides localized function updates through three stages:
(i) \textbf{Contrastive Trajectory Sampling and Analysis}, which identifies behavioral deficiencies from execution evidence;
(ii) \textbf{Module-wise Harness Evolution}, which converts the findings into localized function updates; and
(iii) \textbf{Validation Gates}, which screen modifications for executability and generalizability.

The agent under evolution serves as the \textbf{Code-Modify Agent} for both trajectory analysis and harness modification, processing evidence in batches and restricting edits to the target module.
The independent lineages can be processed in parallel.
After module-wise evolution, \textbf{Cross-Module Integration} combines the evolved modules and resolves potential conflicts.
Throughout evolution, \textbf{Function Merge} controls library redundancy, while a \textbf{Task-Aware Function Composer} selects task-appropriate functions from each module.









\subsection{Harness Modularization}
\label{sec:harness_modularization}

We reorganize the behavioral mechanisms of Terminus-2 in Harbor~\citep{merrill2026terminalbenchbenchmarkingagentshard} into five functional modules.
Based on our analysis of harness implementations and execution trajectories, the decomposition covers mechanisms that directly mediate agent--environment interaction.
Infrastructure components, such as sandbox initialization, parallel execution, and LLM communication, remain outside the evolution scope.

\textbf{(1) Agent Loop} controls the reasoning--action--observation cycle, including execution flow, interaction control, and recovery behavior;
\textbf{(2) Observation Management} preserves task-relevant environmental feedback while filtering or compressing noisy observations;
\textbf{(3) Tool Use} manages tool selection, invocation, and validation;
\textbf{(4) Context Management} maintains interaction history through information retention, compression, and retrieval; and
\textbf{(5) Task Completion Detection} determines whether the task is complete or further interaction is required.

Each module defines a distinct functional scope for subsequent evolution.
During execution, the Agent Loop coordinates the other modules and constructs prompts for the LLM to interact with the environment.
Module interactions and function interfaces are detailed in Appendix~\ref{appendix:harness_architecture}.

\subsection{Contrastive Trajectory Sampling and Analysis}
\label{Sec:Contrastive_Trajectory_Analysis}


For each evolution instance $x_i$, we collect $K$ trajectories $(tra_i^1,\ldots,tra_i^K)$.
The task-specific evaluator assigns each trajectory a binary reward $r_i^k \in \{0,1\}$, where $1$ indicates success.
Based on these rewards, we categorize tasks as \textbf{Positive} if all $K$ rollouts succeed, \textbf{Contrastive} if some succeed and others fail, and \textbf{Negative} if all fail.

A \textbf{Trajectory Memory} stores historical trajectories and rewards for each task across evolution epochs, providing additional evidence when current rollouts are insufficient.

The Code-Modify Agent analyzes each task with respect to the target module.
For the \textbf{Contrastive group}, it compares successful and failed trajectories of the same task to identify function-level factors associated with different outcomes.
For the \textbf{Negative group}, it retrieves a historical success, when available, and pairs it with a current failed trajectory for same-task comparison.
Otherwise, it diagnoses evident execution deficiencies from failed trajectories alone.
For the \textbf{Positive group}, it examines successful trajectories for opportunities to improve execution quality and efficiency.

The diagnoses are recorded as structured JSON findings specifying the module under analysis, supporting trajectory evidence, the rationale for or against modification, and a proposed change when applicable.
The analysis prompt and output schema are provided in Appendices~\ref{appendix:Trajectory_analysis} and~\ref{appendix:Analysis_Finding_Format}.

\subsection{Module-wise Harness Modification}
\label{sebsection_harness_evolution}

Within each independent evolution lineage, the Code-Modify Agent converts the findings from Sec.~\ref{Sec:Contrastive_Trajectory_Analysis} into function updates restricted to the target module.
Two mechanisms guide the selection and implementation of these updates.

\noindent\textbf{Modification Target Selection.}~~
For each batch, we consolidate semantically similar diagnoses targeting the same function into candidate modifications. Each candidate receives a vote count equal to the number of distinct tasks providing supporting evidence. We then rank the candidates by vote count and prioritize the top-\(K\) candidates for modification, thereby focusing on recurring deficiencies rather than instance-specific failures. We set \(K=10\) in all experiments.

\noindent\textbf{Evolution History.}~~
For each function, we record previous code changes and the functionality introduced by each revision.
The Code-Modify Agent uses this history to preserve accumulated capabilities and avoid repeated or contradictory modifications across iterations.
The modification prompt is provided in Appendix~\ref{appendix:harness_modification}.

\subsection{Validation Gates}
\label{subsec:validation_gates}

Each function update must pass three sequential validation gates before being retained.
A failure at any gate triggers rollback using the recorded diffs.

\noindent\textbf{Program Check.}~~
Static checks verify the modified harness's validity and compatibility with the surrounding system, including imports and protocol compliance.

\noindent\textbf{Diff Review.}~~
The Code-Modify Agent inspects the modification diff for task-specific solutions, heuristics, or conditions unlikely to generalize beyond the evolution instances.
The review prompt is provided in Appendix~\ref{appendix:diff_review}.

\noindent\textbf{Execution Validation.}~~
We randomly sample two tasks from the current batch and execute them using the modified harness to detect runtime errors or violations of the expected execution protocol.

Additional validation details are provided in Appendix~\ref{appendix:validation_details}.

\subsection{Cross-Module Integration}
\label{subsec:cross_module_integration}

After independent evolution, we combine the validated module variants into a unified harness.
Although each variant is developed within its own functional scope, composing them may introduce duplicated mechanisms, conflicting behaviors, or inconsistent interactions.

We therefore perform one additional \textbf{cross-module integration epoch} on the evolution set.
The Code-Modify Agent analyzes trajectories from the integrated harness to identify cross-module conflicts, then refines module interactions by removing duplicated mechanisms, clarifying responsibilities, and adjusting coordination logic.
The integration prompt is provided in Appendix~\ref{appendix:merge_mechanism}.

\subsection{Function Library Management}
\label{subsec:function_composition}

As module-wise evolution expands the function library, we control within-module redundancy and task-specific function activation through two complementary mechanisms.

\noindent\textbf{Function Merge.}~~
During evolution, an LLM compares function descriptions and behaviors within each module, merging functions with similar or overlapping functionality to reduce redundancy while preserving learned capabilities.
Function Merge is not applied during downstream evaluation.

\noindent\textbf{Task-Aware Function Composition.}~~
Each evolved function has a natural-language description of its current behavior.
Given the task description and descriptions of all candidate functions, an LLM selects a task-relevant subset, and only these functions are activated during execution.
The composer operates during both trajectory collection and downstream evaluation.

After cross-module integration, the function library is frozen.
Downstream evaluation permits task-aware selection from this fixed library, but no further function merging or modification.

\section{Evolution Dataset and Protocol}
\label{sec:evolution_dataset_and_protocol}

Guided by high-level task categories from the \textbf{TerminalBench} and \textbf{SWE-Bench} families, human annotators retrieve publicly available resources from external sources, including
\href{https://github.com/}{GitHub repositories},
\href{https://huggingface.co/}{Hugging Face},
\href{https://www.kaggle.com/}{Kaggle},
and \href{https://docs.kernel.org/}{Linux kernel documentation}.
These resources are used to construct new executable tasks in the Harbor format.
Benchmarks provide only category-level retrieval guidance; no benchmark instances or task-specific information are used during task construction.

All instances undergo a multi-stage quality-control pipeline.
LLM-based filtering assesses \textbf{Environment Completeness}, checking the availability of required data, dependencies, files, and tools;
\textbf{Practicality and Non-triviality}, excluding artificial or overly simple tasks with limited reasoning or interaction; and
\textbf{Evaluator Validity}, requiring tests to assess functional correctness rather than superficial completion.
Executable validation requires the reference \texttt{solution.sh} to pass all task-specific tests and a no-op submission to receive no positive reward.
Each instance is then manually reviewed for task clarity, environment completeness, and evaluator correctness, followed by LLM-based semantic similarity screening to remove instances with high overlap with benchmarks.

As shown in Fig.~\ref{fig:evolution_data_distribution} in Appendix~\ref{app:the_detail_of_curated_data}, we curate 2,000 high-quality, benchmark-disjoint evolution instances, with relatively balanced task categories and broadly comparable distributions between the TB-related and SWE-related subsets.
Their distinct construction sources limit overlap: SWE-related tasks involve repository-level software engineering on GitHub, whereas TB-related tasks focus on direct terminal interaction.
The dataset provides a foundation for studying whether harness self-evolution yields generalizable improvements rather than benchmark-specific adaptations.

\section{Experiment Setting}
\label{sec:exp_settings}

\subsection{Benchmarks}
To validate the effectiveness of ModularRSI, we evaluate it against the baseline on two recent and challenging terminal-related benchmarks: \textbf{TerminalBench 2.0}~\citep{merrill2026terminalbenchbenchmarkingagentshard} and \textbf{SWE-Bench-Verified}~\citep{jimenez2024swe}. TerminalBench 2.0 contains \textbf{89} diverse long-horizon terminal tasks, while SWE-Bench-Verified contains \textbf{500} human-validated software engineering tasks from real-world repositories. Together, they provide complementary evaluation of general terminal interaction and repository-level software engineering. For reproducibility, we use the original released benchmark versions and conduct all evaluations under the Harbor framework.

\subsection{Metrics}
Considering the stability of the evaluation results and the need to compare both the efficiency and performance of different harnesses, we evaluate the following four metrics:

\noindent\textbf{Accuracy (Acc):}~~
Accuracy measures the average success rate over all rollout trajectories. A trajectory is considered successful if the agent successfully completes the corresponding task according to the task-specific evaluator.

\noindent\textbf{Pass@3:}~~Pass@3 measures whether the agent can successfully solve a task within three independent rollout attempts. A task is counted as solved if at least one of the three trajectories succeeds, and the final score is averaged across all evaluation tasks.

\noindent\textbf{StepNum:}~~
StepNum measures the average number of agent-environment interaction steps required per rollout. It reflects the execution efficiency of the agent, with fewer steps generally indicating a more efficient problem-solving process.

\noindent\textbf{$\text{Pass}^3$:}~~
$\text{Pass}^3$ measures the consistency and reliability of the agent across $3$ independent rollout attempts. A task is counted as successfully solved only if all $3$ trajectories succeed, and the final score is averaged across all evaluation tasks.

\subsection{Implementation Details}
\label{subsec:imp_details}

Each module-specific evolution lineage in ModularRSI is evolved for \textbf{3 epochs}. The five independently evolved modules are subsequently combined through one cross-module integration epoch. 
For the experiments with different methods or settings that evolve the entire harness jointly, we adjust the number of epochs to match the aggregate evolution budget of ModularRSI, as detailed in Sec.~\ref{subsec:effect_modular_evolution} and Sec.~\ref{subsec:Comparing_with_Existing_RSI_Methods}.
Due to computational and time constraints, we select \textbf{120} instances from each of the TB-related and SWE-related subsets of the 2,000-instance evolution dataset for harness evolution. The primary model used for evolution is \textbf{DeepSeek-V4-Flash-Preview}. We provision the deployed models with a \textbf{TPM limit of 2 million tokens} and use a \textbf{batch size of 10} throughout the evolution process.
In this work, we regard Terminus-2 as the No-Evolution (No-Evol) Baseline.



\section{Results}

\subsection{Generalization Beyond Evolution Experience}

The central challenge of Harness RSI is not only improving performance on the tasks used for evolution, but discovering harness modifications that generalize beyond the evolution experience. To evaluate this capability, we construct two independent evolution sets, i.e., TB-related and SWE-related. From each set, we sample 120 instances for harness evolution, resulting in 120 TB-related and 120 SWE-related evolution tasks. We then evaluate the evolved harnesses on both TerminalBench 2.0 and SWE-Bench-Verified under in-domain and out-of-domain settings. 
The evolution sets contain no downstream benchmark instances, trajectories, evaluator feedback, or task-specific content; benchmark taxonomies are used only as coarse category-level guidance for source coverage.

\begin{table}[t]
\centering
\caption{
Cross-benchmark transfer of evolved harnesses.
After evolution on TB-related or SWE-related tasks, each harness is evaluated on both held-out benchmarks.
}
\label{tab:cross_domain_generalization}

\footnotesize
\setlength{\tabcolsep}{4pt}
\begin{tabular}{lcccc|cccc}
\toprule
& \multicolumn{4}{c|}{\textbf{SWE-Bench-Verified}}
& \multicolumn{4}{c}{\textbf{TerminalBench 2.0}} \\
\cmidrule(lr){2-5}
\cmidrule(lr){6-9}

\textbf{Evolution Set}
& \textbf{Eval. Setting}
& \textbf{Acc $\uparrow$}
& \textbf{Pass@3 $\uparrow$}
& \textbf{Pass$^{3}$ $\uparrow$}
& \textbf{Eval. Setting}
& \textbf{Acc $\uparrow$}
& \textbf{Pass@3 $\uparrow$}
& \textbf{Pass$^{3}$ $\uparrow$} \\
\midrule

No Evolution
& -- & 73.40 & 83.20 & 62.80
& -- & 47.57 & 58.43 & 30.34 \\

TB-Related
& Out-of-Domain & 75.80 & 84.60 & 66.20
& In-Domain & \textbf{52.43} & \textbf{65.17} & \textbf{35.96} \\

SWE-Related
& In-Domain & \textbf{76.40} & \textbf{85.40} & \textbf{66.80}
& Out-of-Domain & 49.44 & 60.67 & 30.34 \\

\bottomrule
\end{tabular}
\end{table}

As shown in Tab.~\ref{tab:cross_domain_generalization}, with \textbf{DeepSeek-V4-Flash-Preview} as the backbone model, ModularRSI consistently improves performance across both evaluation benchmarks under different evolution settings. In the in-domain setting, the evolved harness improves accuracy from 47.57 to 52.43 on TerminalBench 2.0 and from 73.40 to 76.40 on SWE-Bench-Verified. We further analyze performance across successive RSI generations and observe a largely monotonic improvement throughout the evolution process (see Appendix~\ref{Appendix: Evolution Curve}).

More importantly, these gains transfer beyond the evolution domain. The harness evolved on TB-related data improves SWE-Bench-Verified accuracy to 75.80, while the harness evolved on SWE-related data improves TerminalBench 2.0 accuracy to 49.44. These results suggest that ModularRSI discovers reusable improvements to the underlying agent execution mechanism, rather than merely specializing to the tasks encountered during evolution.

Notably, the gain is particularly pronounced for $\mathrm{Pass}^{3}$ on TB 2.0, which increases from 30.34 to 35.96. This improvement suggests that harness evolution not only increases average task-solving capability but also enhances execution reliability, reducing stochastic failures across repeated trials.

\subsection{Cross-Model Generalization}

\begin{wraptable}{r}{0.55\linewidth}
\vspace{-8pt}
\centering
\footnotesize
\setlength{\tabcolsep}{4pt}

\caption{Generalization across different inference models with an evolved harness on TB 2.0.}
\label{tab:model_generalization}

\resizebox{\linewidth}{!}{
\begin{tabular}{llccc}
\toprule
\textbf{Infer Model} 
& \textbf{Method} 
& \textbf{Acc $\uparrow$} 
& \textbf{Pass@3 $\uparrow$} 
& \textbf{Pass$^{3}$ $\uparrow$} \\
\midrule

\multirow{2}{*}{GLM-5.2} 
& No-Evol Baseline 
& 59.55 
& 70.79 
& 46.07 \\

& ModularRSI 
& \textbf{61.80} 
& \textbf{74.16} 
& \textbf{49.44} \\

\midrule

\multirow{2}{*}{MiniMax-2.5} 
& No-Evol Baseline 
& 41.57 
& 56.18 
& 24.72 \\

& ModularRSI 
& \textbf{44.94} 
& \textbf{57.30} 
& \textbf{30.34} \\

\midrule

\multirow{2}{*}{DS-V4-Flash} 
& No-Evol Baseline 
& 47.57 
& 58.43 
& 30.34 \\

& ModularRSI
& \textbf{52.43} 
& \textbf{65.17} 
& \textbf{35.96} \\

\bottomrule
\end{tabular}
}

\vspace{-8pt}
\end{wraptable}

Beyond task-level generalization, we further examine whether the discovered harness improvements are transferable across different foundation models. Since different models exhibit different behaviors in planning, tool interaction, and error recovery, a model-specific optimization may fail to generalize.

To evaluate this capability, we freeze the harness evolved with DeepSeek-V4-Flash Preview on the TB-related evolution set and apply it to different foundation models on TerminalBench 2.0.

As shown in Tab.~\ref{tab:model_generalization}, the evolved harness consistently improves Acc, Pass@3, and Pass$^{3}$ across all evaluated foundation models. For example, the evolved harness improves GLM-5.2 accuracy from 59.55 to 61.80 and MiniMax-2.5 accuracy from 41.57 to 44.94. 
Furthermore, we apply ModularRSI with different models as the evolution backbone. As shown in Appendix~\ref{appendix:model_generation}, ModularRSI consistently improves harness performance across models with varying capabilities, although the magnitude of improvement differs across models.

These results demonstrate that ModularRSI learns transferable improvements to the agent execution process rather than exploiting model-specific behaviors.

\subsection{Effect of Modular Evolution}
\label{subsec:effect_modular_evolution}



We examine whether the gains of ModularRSI arise from greater evolution exposure or from modular evolution itself. We compare against joint all-module and non-modular whole-harness evolution. All methods use 3 rollout trajectories per task, and both alternatives are evolved for 16 epochs, matching the aggregate three-epoch evolution of five modules followed by one cross-module integration epoch,  to match the aggregate evolution budget of ModularRSI.

\begin{wraptable}{r}{0.63\linewidth}
\vspace{-16pt}
\centering
\footnotesize
\setlength{\tabcolsep}{4pt}

\caption{Comparison of different harness evolution strategies on TerminalBench 2.0.}
\label{tab:joint_vs_merge}

\resizebox{\linewidth}{!}{
\begin{tabular}{lcccc}
\toprule
\textbf{Method} &
\textbf{Acc $\uparrow$} &
\textbf{Pass@3 $\uparrow$} &
\textbf{Pass$^{3}$ $\uparrow$} &
\textbf{StepNum $\downarrow$} \\
\midrule

No-Evol Baseline
& 47.57 & 58.43 & 30.34 & 34.70 \\

Non-modular Evolution
& 46.44 & 64.04 & 24.72 & \textbf{29.03} \\

Joint All-Module Evolution
& 44.19 & 61.80 & 24.72 & 44.34 \\

ModularRSI (Ours)
& \textbf{52.43}
& \textbf{65.17}
& \textbf{35.96}
& 35.57 \\

\bottomrule
\end{tabular}
}

\vspace{-8pt}
\end{wraptable}

As shown in Tab.~\ref{tab:joint_vs_merge}, a larger self-modification space is not necessarily beneficial for RSI. Under the same trajectory and evolution budget, both joint all-module evolution and unrestricted whole-harness evolution perform worse than the initial harness, whereas independently localized evolution improves accuracy from 47.57 to 52.43. This suggests that, in benchmark-disjoint harness RSI, \textbf{the ability to modify more of the system can increase credit-assignment ambiguity rather than improve adaptation.}

\begin{wraptable}{r}{0.62\linewidth}
\vspace{-8pt}
\centering
\footnotesize
\setlength{\tabcolsep}{3pt}

\caption{
Performance of single-module evolution and cross-module integration on TB 2.0.
Each row under \textit{Single-Module Evolution} reports the harness obtained by independently evolving one module from modularized Terminus-2.
}
\label{tab:single_module_evolution}

\resizebox{\linewidth}{!}{
\begin{tabular}{lcccc}
\toprule
\textbf{Method} &
\textbf{Acc $\uparrow$} &
\textbf{Pass@3 $\uparrow$} &
\textbf{Pass$^{3}$ $\uparrow$} &
\textbf{StepNum $\downarrow$} \\
\midrule

No-Evol Baseline
& 47.57 & 58.43 & 30.34 & 34.70 \\

\midrule
\multicolumn{5}{l}{\textit{Single-Module Evolution}} \\

Context Management
& 49.44 & 61.80 & 30.34 & 35.10 \\

Tool Use
& 50.19 & 62.92 & 30.34 & 41.28 \\

Agent Loop
& 50.56 & 64.04 & 34.83 & 40.40 \\

Observation Management
& 49.81 & \textbf{65.17} & 33.70 & \textbf{22.50} \\

Task Completion Detection
& 49.44 & \textbf{65.17} & 31.46 & 31.06 \\

\midrule
\multicolumn{5}{l}{\textit{Cross-Module Integration}} \\

ModularRSI (Ours)
& \textbf{52.43}
& \textbf{65.17}
& \textbf{35.96}
& 35.57 \\

\bottomrule
\end{tabular}
}

\vspace{-42pt}
\end{wraptable}

Tab.~\ref{tab:single_module_evolution} further shows that all independently evolved modules improve accuracy, but with different behavioral effects. Agent Loop provides the largest single-module accuracy gain, while Observation Management substantially reduces execution steps from 34.70 to 22.50. Cross-module integration further improves accuracy to 52.43, indicating that independently evolved modules capture complementary improvements that can be effectively combined.

\subsection{The Ablation Study}

\begin{wraptable}{r}{0.51\linewidth}
\vspace{-22pt}
\centering
\small
\setlength{\tabcolsep}{4pt}

\caption{Ablation study of CTA, FLM, and VG in ModularRSI on TB 2.0.}
\label{tab:contrastive_analysis_ablation}

\resizebox{\linewidth}{!}{
\begin{tabular}{lccc}
\toprule
\textbf{Method} &
\textbf{Acc $\uparrow$} &
\textbf{Pass@3 $\uparrow$} &
\textbf{Pass$^{3}$ $\uparrow$} \\
\midrule

ModularRSI
& \textbf{52.43}
& \textbf{65.17}
& \textbf{35.96} \\

ModularRSI w/o CTA
& 50.56
& 64.04
& 34.83 \\

ModularRSI w/o FLM
& 49.81
& 62.92
& 31.46 \\

ModularRSI w/o VG
& 48.31
& 61.79
& 30.34 \\

\hdashline

No-Evol Baseline
& 47.57
& 58.43
& 30.34 \\

\bottomrule
\end{tabular}
}

\vspace{-8pt}
\end{wraptable}

We conduct ablation studies to examine the contributions of three key components in ModularRSI: \textbf{Contrastive Trajectory Analysis (CTA)}, \textbf{Function Library Management (FLM)}, and \textbf{Validation Gates (VG)}. We remove each component individually while keeping the remaining evolution setting unchanged, and evaluate the resulting harnesses on TerminalBench 2.0.

As shown in Tab.~\ref{tab:contrastive_analysis_ablation}, all three ablations lead to clear performance degradation, suggesting that the components address complementary aspects of harness evolution. Removing CTA decreases accuracy from 52.43 to 50.56, indicating that contrasting successful and failed trajectories provides a more reliable signal for identifying recurring behavioral deficiencies. Without VG, accuracy drops to 48.31, suggesting that explicit validation is important for preventing ineffective or harmful modifications from being retained during evolution. FLM has a substantial impact: removing it reduces accuracy to 49.81, with further degradation observed in Pass@3 and Pass$^3$. This result suggests that simply accumulating evolved functions is insufficient; consolidating redundant functions and selecting task-relevant ones are important for reducing interference among independently evolved improvements.

The case studies in Appendix~\ref{appendix:trajectory_group_cases} further illustrate how Contrastive Trajectory Analysis identifies recurring behavioral deficiencies across trajectories and converts them into reusable harness improvements rather than task-specific fixes.

\subsection{Comparing with Existing RSI Methods}
\label{subsec:Comparing_with_Existing_RSI_Methods}










We compare ModularRSI with AHE and Meta-Harness under the unified benchmark-disjoint evolution protocol described in Sec.~\ref{subsec:imp_details}. All methods start from the same Terminus-2 harness and are evolved using the same evolution data, backbone model (DeepSeek-V4-Flash-preview), and matched evolution budget.
The implementation details are provided in Appendix~\ref{appendix:Existing_Method_comparision_Settings}

\begin{wraptable}{r}{0.48\linewidth}
\vspace{-8pt}
\centering
\small
\setlength{\tabcolsep}{4pt}

\caption{Comparison with existing harness evolution methods on TerminalBench 2.0.}
\label{tab:comparison_existing_methods}

\resizebox{\linewidth}{!}{
\begin{tabular}{lccc}
\toprule
\textbf{Method} &
\textbf{Acc $\uparrow$} &
\textbf{Pass@3 $\uparrow$} &
\textbf{Pass$^{3}$ $\uparrow$} \\
\midrule

No-Evol Baseline
& 47.57 & 58.43 & 30.34 \\

Meta-Harness
& 49.44 & 61.80 & 31.40 \\

AHE

& 49.81 & 64.04 & 33.70 \\ \hdashline
ModularRSI (Ours)
& \textbf{52.43}
& \textbf{65.17}
& \textbf{35.96} \\

\bottomrule
\end{tabular}
}

\vspace{-8pt}
\end{wraptable}


As shown in Tab.~\ref{tab:comparison_existing_methods}, AHE and Meta-Harness provide only modest improvements over the initial harness, increasing accuracy from 47.57 to 49.81 and 49.44, respectively. In contrast, ModularRSI achieves 52.43 accuracy, with consistent improvements also observed in Pass@3 and Pass$^3$. Since all compared methods share the same initial harness, evolution data, foundation model, and evolution budget, the larger improvement cannot be attributed simply to greater exposure to trajectories or additional evolution rounds. Instead, the results suggest that modular evolution with more precise credit assignment enables more effective discovery of generalizable harness improvements.

This result is consistent with our central design hypothesis: under benchmark-disjoint evolution, effective harness RSI requires converting execution experience into reusable mechanism-level improvements rather than broadly modifying the harness from task-level feedback. ModularRSI constrains this process through contrastive trajectory attribution and module-restricted evolution, which narrows the credit-assignment space and reduces interference between unrelated harness mechanisms. The results therefore suggest that these constraints are particularly beneficial when evolution experience and downstream evaluation tasks are disjoint.

\subsection{Scaling with Evolution Data}

\begin{wrapfigure}{r}{0.4\linewidth}
    \centering
    \vspace{-22pt}
    \includegraphics[width=0.96\linewidth]{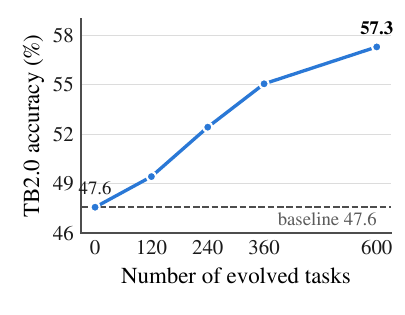}
    \caption{Accuracy of ModularRSI on TerminalBench 2.0 as a function of evolution dataset size.}
    \label{fig:tb2_scaling}
    \vspace{-8pt}
\end{wrapfigure}

Benchmark-disjoint evolution data and fair comparisons under a unified experimental protocol are important for assessing the generalization of RSI methods. To assess the quality and utility of our curated evolution dataset, we examine whether increasing the number of evolution instances yields further performance gains for ModularRSI, while keeping the number of evolution epochs fixed at one. We use DeepSeek-V4-Flash-Preview as the backbone model and evaluate the evolved harnesses on TerminalBench 2.0.

As shown in the Fig.~\ref{fig:tb2_scaling}, we run ModularRSI with 120, 240, 360, and 600 evolution instances. Performance improves consistently as the amount of evolution data increases, exhibiting a sustained upward trend. These results demonstrate that ModularRSI can benefit from scaling up high-quality evolution data and provide further evidence of the quality and utility of our curated dataset, suggesting that a broader subset of its instances can effectively support harness evolution.

\section{Conclusion}

We present \textbf{ModularRSI}, a benchmark-disjoint and modular framework for generalizable harness self-improvement. By contrasting execution trajectories and evolving harness modules independently, ModularRSI identifies reusable mechanism-level improvements while reducing task-specific adaptation. Experiments on TerminalBench 2.0 and SWE-Bench Verified demonstrate consistent gains across unseen tasks, domains, and foundation models.
Overall, our results demonstrate the potential of benchmark-disjoint, contrastive, and modular evolution for building more generalizable self-improving agent harnesses.

\bibliography{iclr2027_conference}
\bibliographystyle{iclr2027_conference}
\clearpage

\appendix

\section{The Comparison of Different Harness RSI Methods}

We systematically compare the current Harness RSI methods in Tab.~\ref{tab:self_evolving_harness_comparison}.

\newcommand{\cmark}{\textcolor{green!60!black}{\ding{51}}}
\newcommand{\xmark}{\textcolor{red!70!black}{\ding{55}}}

\begin{table*}[t]
\centering
\footnotesize
\setlength{\tabcolsep}{3.5pt}
\renewcommand{\arraystretch}{1.12}
\caption{
Comparison of representative self-evolving agent and harness frameworks.
Do Not Use Benchmark Data indicates whether benchmark instances, trajectories,
or benchmark-derived rewards are excluded from harness evolution.
}
\begin{tabular}{lccc}
\toprule

\textbf{Method}
&
\makecell[c]{\textbf{Modular}\\\textbf{Evolution}}
&
\makecell[c]{\textbf{Whole-Harness}\\\textbf{Evolution}}
&
\makecell[c]{\textbf{Not Use}\\\textbf{Benchmark Data}}
\\

\midrule

AutoHarness~\citep{lou2026autoharness}
& \xmark & \xmark & \xmark \\

Meta-Harness~\citep{lee2026meta}
& \xmark & \cmark & \xmark \\

AHE~\citep{lin2026agentic}
& \xmark & \xmark & \xmark \\

Self-Harness~\citep{zhang2026self}
& \xmark & \xmark & \xmark \\

RHO~\citep{pan2026evolvingagentsdarkretrospective}
& \xmark & \xmark & \xmark \\

HarnessFix~\citep{chen2026failed}
& \cmark & \xmark & \xmark \\

Living-Harness~\citep{du2026livingharnessinteractiveagentevolver}
& \cmark & \xmark & \xmark \\

HarnessForge~\citep{chen2026harnessforgejointharnesspolicy}
& \cmark & \cmark & \xmark \\


RSEA~\citep{nguyen2026recursiveselfevolvingagentsheldout}
& \cmark & \xmark & \xmark \\

TACO~\citep{ren2026selfevolvingframeworkefficientterminal}
& \xmark & \xmark & \xmark \\

\midrule

\textbf{ModularRSI (Ours)}
& \cmark
& \cmark
& \cmark \\

\bottomrule
\end{tabular}

\label{tab:self_evolving_harness_comparison}

\end{table*}

\section{Harness Architecture}
\label{appendix:harness_architecture}

\begin{figure}[h]
    \centering
    \includegraphics[width=\linewidth]{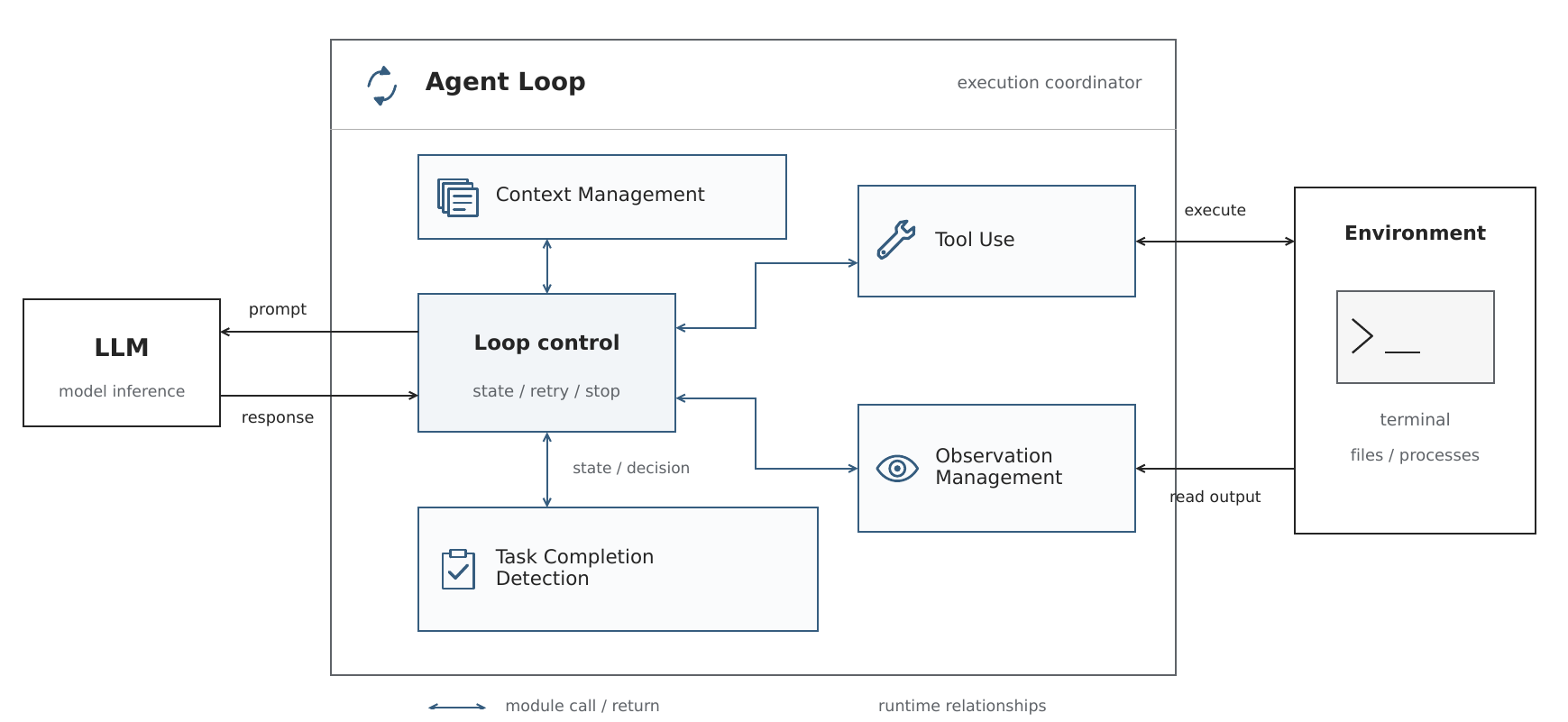}
    \caption{Architecture of the five-module harness.}
    \label{fig:harness_architecture}
\end{figure}

Figure~\ref{fig:harness_architecture} shows how the five modules interact during task execution. The outer box is Agent Loop, and the central \textit{Loop control} box represents its runtime logic: maintaining execution state, scheduling calls, handling retries, and deciding whether to continue. Each of the other four modules exchanges inputs and outputs with this controller. The bidirectional module arrows indicate a call and its return. The LLM and environment are external resources: the LLM generates responses, while the environment runs commands and stores files and processes. Before execution, the composer initializes the harness by selecting task-appropriate functions from each module according to the specific task.

In a normal iteration, Agent Loop first asks Context Management to update the conversation history, compressing it when needed. It then sends the current prompt to the LLM. Tool Use parses the response into commands and a completion signal, executes the commands in the environment, and returns execution results or errors. Observation Management reads the current terminal output and converts it into agent-readable feedback. This differs from Context Management, which works on the accumulated conversation. Agent Loop combines the feedback with tool results, updates its state, and passes that state to Task Completion Detection.

Task Completion Detection returns a stop recommendation and a reason. For example, the detector requires two consecutive completion declarations. Agent Loop maintains that count and issues the confirmation prompt after the first declaration; the detector checks whether the count meets its criterion. The loop follows the recommendation, while evolved loop variants can override it. If execution continues, the loop prepares the next prompt from the feedback. Parsing errors can instead trigger a retry, and session failures or execution limits can end the run. The figure shows these modules' relationships; the arrows do not prescribe a fixed module-to-module execution sequence.

\subsection{Module Interface Excerpts}

Listing~\ref{lst:harness_module_interfaces} shows the main inputs and outputs of all five modules. The signatures follow the implementation; imports, setup methods, and method bodies are omitted. Each implementation is registered with a name, description, and configuration parameters. The shared \texttt{ModuleCtx} provides task configuration, environment access, and recording services.

\begin{lstlisting}[
  style=promptstyle,
  language=Python,
  caption={Selected interfaces of the five harness modules. Ellipses denote omitted implementations.},
  label={lst:harness_module_interfaces}
]
# 1. Agent Loop
# Input: task, conversation, and the other four modules.
# Output: execution status, final text, and failure tag.
class AgentLoop(Protocol):
    async def run(
        self, initial_prompt: str, original_instruction: str,
        observation: Observation, context_mgmt: ContextMgmt,
        tools: ToolSet, verification: VerificationLoop,
        chat: Chat, ctx: ModuleCtx,
    ) -> AgentLoopResult: ...

# 2. Observation Management
# Reads the environment through ctx.
# Output: observation text and updated observation state.
class Observation(Protocol):
    async def capture(
        self, prev: ObsState, ctx: ModuleCtx,
    ) -> tuple[ObsResult, ObsState]: ...

# 3. Tool Use
# Model text -> commands, completion signal, parsing errors.
# A tool call -> execution result (success, output, error).
class ToolSet(Protocol):
    def parse_llm_response(
        self, response: str,
    ) -> LLMResponseParseResult: ...

    async def execute(
        self, call: ToolCall, ctx: ModuleCtx,
    ) -> ToolResult: ...

# 4. Context Management
# Input: conversation history and the original task.
# Output: updated Chat and an optional handoff prompt.
class ContextMgmt(Protocol):
    async def maybe_compress(
        self, chat: Chat, original_instruction: str,
        ctx: ModuleCtx,
    ) -> CompressResult: ...

    async def force_summarize(
        self, chat: Chat, original_instruction: str,
        ctx: ModuleCtx,
    ) -> CompressResult: ...

# 5. Task Completion Detection
# Input: loop state, including recent feedback and signals.
# Output: a stop recommendation and its reason.
class VerificationLoop(Protocol):
    async def should_terminate(
        self, state: AgentLoopState, ctx: ModuleCtx,
    ) -> tuple[bool, str]: ...
\end{lstlisting}

The completion detector returns a recommendation; Agent Loop maintains the state and controls whether execution continues. Observation Management returns current environment feedback, whereas Context Management updates the accumulated conversation history.

\section{Validation Details}
\label{appendix:validation_details}

The \textbf{Program Check} gate includes AST validation, import checks, protocol compliance checks, discovery-contract verification, and static self-attribute audits.
For \textbf{Execution Validation}, two tasks are randomly sampled from the current evolution batch and executed using the modified harness.
Updates introducing runtime errors or violating the expected execution protocol are rejected.
A failure at any validation gate triggers rollback using the recorded diffs to restore the previous version.

\section{The Performance of ModularRSI under Different Foundation Models}
\label{appendix:model_generation}

\begin{wraptable}{r}{0.6\linewidth}
\vspace{-8pt}
\centering
\footnotesize
\setlength{\tabcolsep}{4pt}

\caption{Generalization across different evolution models to run ModularRSI on TB 2.0.}
\label{tab:evol_model_generalization}

\resizebox{\linewidth}{!}{
\begin{tabular}{llccc}
\toprule
\textbf{Evol Model} 
& \textbf{Method} 
& \textbf{Acc $\uparrow$} 
& \textbf{Pass@3 $\uparrow$} 
& \textbf{Pass$^{3}$ $\uparrow$} \\
\midrule

\multirow{2}{*}{DS-V4-Flash-Preview} 
& No-Evol Baseline 
& 47.57 
& 58.43 
& 30.34 \\

& ModularRSI
& \textbf{52.43} 
& \textbf{65.17} 
& \textbf{35.96} \\

\midrule

\multirow{2}{*}{DS-V4-Flash-0731} 
& No-Evol Baseline 
&  61.79
&  73.03
&   50.56\\

& ModularRSI
&  67.42 
&  78.65 
&   56.18\\

\midrule

\multirow{2}{*}{Qwen3.8-Max} 
& No-Evol Baseline 
&    61.05
&    73.03
&  49.44 \\

& ModularRSI
&   64.79
&   75.28
&  55.06 \\

\bottomrule
\end{tabular}
}

\vspace{-8pt}
\end{wraptable}

To demonstrate that our method can be used by models with different capabilities, we run ModularRSI under different foundation models (i.e., DeepSeek-V4-Flash-Preview, DeepSeek-V4-Flash-0731, and Qwen3.8-Max) on the 120 TB-related evolution data.

As shown in Table~\ref{tab:evol_model_generalization}, After evolution with ModularRSI across different models, the resulting harnesses consistently outperform their corresponding baselines by substantial margins, further demonstrating the generalizability of our method.
.

\clearpage

\section{The Detail of Curated Data}
\label{app:the_detail_of_curated_data}
\begin{figure*}[t]
    \centering
    \includegraphics[width=0.99\textwidth]{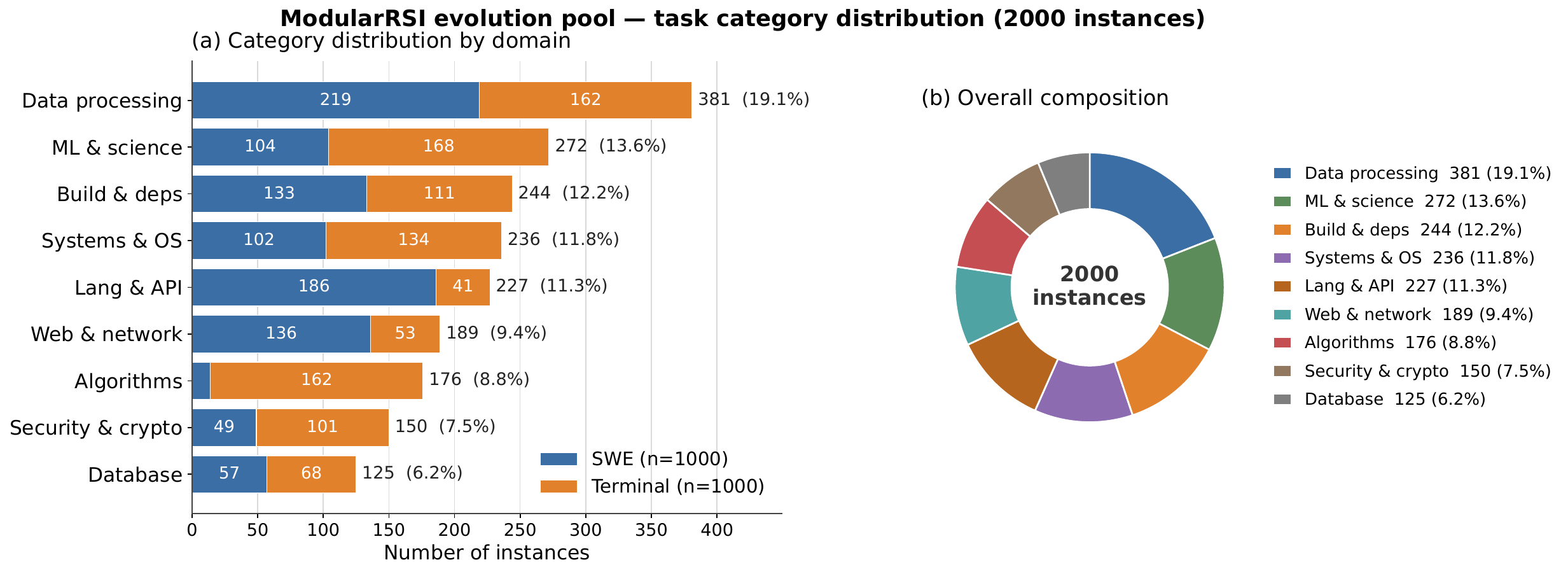}
    \caption{
        The data distribution of our evolution data.
    }
    \label{fig:evolution_data_distribution}
    \vspace{-2mm}
\end{figure*}

We provide the static details of our curated data in Fig.~\ref{fig:evolution_data_distribution}.

\clearpage

\section{The Experiment Settings for Comparing Existing Methods}
\label{appendix:Existing_Method_comparision_Settings}

Considering task compatibility, reproducibility, and method performance, we adapt and reproduce \textbf{AHE} and \textbf{Meta-Harness} under a unified experimental protocol. Specifically, all methods are adapted to the Harbor framework and evolve from the same \textbf{Terminus-2} harness using the same tb-related 120 evolution instances. For all methods, each task is rolled out three times (\(n=3\)), yielding three trajectories per task. ModularRSI evolves each module independently for three epochs, followed by one cross-module integration epoch. To match the aggregate evolution budget, we evolve AHE and Meta-Harness for 16 epochs. For all methods, the harness from the final generation is evaluated on the downstream benchmark. All methods use \textbf{DeepSeek-V4-Flash-Preview} for both evolution and evaluation, with the same TPM limit of 2M tokens per minute. We additionally disable web search for all methods to prevent the evolved harnesses from retrieving task solutions from external sources.

\clearpage

\section{Confidence Interval Analysis of Experimental Results}
\label{appendix:CIAoER}


To assess the statistical reliability of performance differences, we conduct non-parametric paired bootstrap analysis at the task level. For each setting, the fixed harness is independently evaluated three times, and the three binary outcomes of each task are averaged to obtain its task-level success rate.

For each method--baseline comparison, we compute

$$
\Delta \mathrm{Acc}
=
\mathrm{Acc}_{\mathrm{method}}
-
\mathrm{Acc}_{\mathrm{baseline}}.
$$

We perform 100,000 paired bootstrap resamples over tasks, using identical sampled task indices for both methods while retaining the three runs of each task together. The 2.5th and 97.5th percentiles of the bootstrap distribution are reported as the 95

\begin{table*}[t]
\centering
\footnotesize
\setlength{\tabcolsep}{5pt}

\caption{Paired bootstrap analysis of the main experimental results. Accuracy differences are computed relative to the corresponding baseline, with 95\% confidence intervals estimated by task-level paired bootstrap resampling.}
\textbf{(a) Cross-benchmark generalization}

\vspace{2pt}

\begin{tabular}{llccc}
\toprule
\textbf{Evolution Set} &
\textbf{Evaluation Benchmark} &
\textbf{Acc. $\uparrow$} &
\textbf{$\Delta$ Acc.} &
\textbf{95\% CI of $\Delta$} \\
\midrule
No Evolution & SWE-Bench Verified & 73.40 & -- & -- \\
TB-Related   & SWE-Bench Verified & 75.80 & +2.40 & [0.73, 3.33] \\
SWE-Related  & SWE-Bench Verified & 76.40 & +3.00 & [1.06, 4.26] \\
\midrule
No Evolution & TerminalBench 2.0 & 47.57 & -- & -- \\
TB-Related   & TerminalBench 2.0 & 52.43 & +4.86 & [3.37, 9.36] \\
SWE-Related  & TerminalBench 2.0 & 49.44 & +1.87 & [1.49, 5.24] \\
\bottomrule
\end{tabular}

\vspace{8pt}

\textbf{(b) Cross-model generalization on TerminalBench 2.0}

\vspace{2pt}

\begin{tabular}{llccc}
\toprule
\textbf{Inference Model} &
\textbf{Method} &
\textbf{Acc. $\uparrow$} &
\textbf{$\Delta$ Acc.} &
\textbf{95\% CI of $\Delta$} \\
\midrule
GLM-5.2 & Baseline   & 59.55 & -- & -- \\
GLM-5.2 & ModularRSI & 61.80 & +2.25 & [1.12, 5.24] \\
\midrule
MiniMax-2.5 & Baseline   & 41.57 & -- & -- \\
MiniMax-2.5 & ModularRSI & 44.94 & +3.37 & [2.25, 7.49] \\
\midrule
DeepSeek-V4-Flash & Baseline   & 47.57 & -- & -- \\
DeepSeek-V4-Flash & ModularRSI & 52.43 & +4.86 & [3.37, 9.36] \\
\bottomrule
\end{tabular}

\vspace{8pt}

\textbf{(c) Comparison with existing harness evolution methods}

\vspace{2pt}

\begin{tabular}{lccc}
\toprule
\textbf{Method} &
\textbf{Acc. $\uparrow$} &
\textbf{$\Delta$ Acc.} &
\textbf{95\% CI of $\Delta$} \\
\midrule
Baseline          & 47.57 & -- & -- \\
Meta-Harness      & 49.44 & +1.87 & [-0.07, 2.24] \\
AHE               & 49.81 & +2.24 & [-0.03, 2.62] \\
ModularRSI (Ours) & \textbf{52.43} & +4.86 & [3.37, 9.36] \\
\bottomrule
\end{tabular}

\vspace{8pt}

\textbf{(d) Comparison of different harness evolution strategies}

\vspace{2pt}

\begin{tabular}{lccc}
\toprule
\textbf{Method} &
\textbf{Acc. $\uparrow$} &
\textbf{$\Delta$ Acc.} &
\textbf{95\% CI of $\Delta$} \\
\midrule
Baseline                   & 47.57 & -- & -- \\
Non-modular Evolution      & 46.44 & -1.13 & [-2.24, -0.74] \\
Joint All-Module Evolution & 44.19 & -3.38 & [-3.74, -1.50] \\
ModularRSI (Ours)          & \textbf{52.43} & +4.86 & [3.37, 9.36] \\
\bottomrule
\end{tabular}

\label{tab:confidence_interval_analysis}

\end{table*}

\clearpage

\section{Effect of Evolution Data Quality}

\begin{wrapfigure}{r}{0.48\linewidth}
    \centering
    \vspace{-8pt}
    \includegraphics[width=0.96\linewidth]{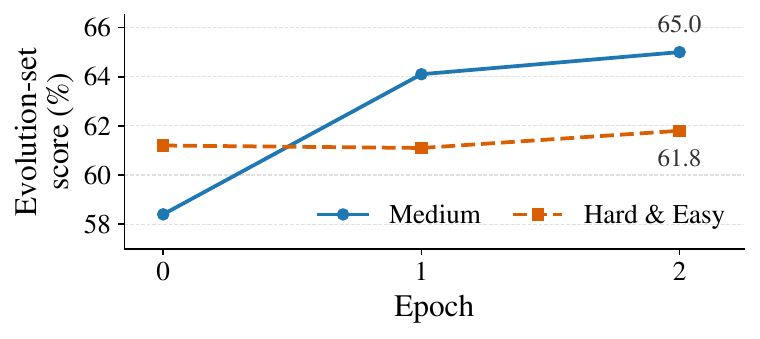}
    \caption{Accuracy variation during evolution under different difficulty distributions. DeepSeek-V4-Flash Preview is used as the foundation model.}
    \label{fig:evolution_difficulty}
    \vspace{-8pt}
\end{wrapfigure}

Beyond the evolution strategy, the quality of evolution experience also affects the effectiveness of Harness RSI. Since harness evolution relies on execution feedback to identify systematic failures and improve interaction strategies, the informativeness of evolution tasks is crucial. In particular, the difficulty distribution of evolution data plays an important role in determining the quality of evolution signals. Moderately difficult tasks are more likely to provide informative contrasts: easy tasks contain insufficient failure cases, while extremely difficult tasks rarely yield successful trajectories that reveal effective interaction strategies. Therefore, we hypothesize that a balanced difficulty distribution can facilitate the discovery of generalizable harness improvements.

To verify this hypothesis, we construct two evolution settings with different difficulty distributions: a \textit{Medium-centered} setting and a \textit{Hard \& Easy} setting. 
The detailed procedure for constructing these evolution datasets is described in the Appendix.~\ref{appendix:difficulty_setting}.

\begin{wraptable}{r}{0.35\linewidth}
\vspace{-12pt}
\centering
\footnotesize
\caption{Performance of Different Evolution-Data Distributions on SWE-Bench Verified.}
\label{tab:difficulty_performance}

\resizebox{0.95\linewidth}{!}{
\begin{tabular}{lc}
\toprule
\textbf{Difficulty Distribution} & \textbf{Acc $\uparrow$} \\
\midrule
Medium-centered & \textbf{76.40} \\
Hard \& Easy & 74.20 \\
\bottomrule
\end{tabular}
}

\vspace{-8pt}
\end{wraptable}

As shown in Fig.~\ref{fig:evolution_difficulty}, the Medium-centered setting achieves consistently stronger improvement throughout the evolution process. 
Moreover, as shown in Tab.~\ref{tab:difficulty_performance} the evolved harness generalizes better to downstream evaluation: the harness evolved with the Medium-centered distribution achieves 76.40\% accuracy on SWE-Bench Verified, outperforming the Hard \& Easy setting by 2.20 percentage points (74.20\%). This demonstrates that evolution data with more informative trajectory contrasts leads to more effective and generalizable harness improvements.

\section{Structured Analysis Findings}
\label{appendix:Analysis_Finding_Format}

The three trajectory groups in Sec.~\ref{Sec:Contrastive_Trajectory_Analysis}
describe the available reward evidence. The implementation refines these
groups into routing buckets. The buckets select an analysis path. They do not
change the reward of a trajectory. Table~\ref{tab:routing_mapping} shows the
mapping. Tasks with infrastructure-only failures are excluded from harness
diagnosis.

\begin{table}[h]
\centering
\small
\setlength{\tabcolsep}{4pt}
\renewcommand{\arraystretch}{1.12}
\caption{Mapping from the method-level trajectory groups to the routing buckets used by the implementation.}
\begin{tabular}{lp{0.60\linewidth}}
\toprule
\textbf{Method group} & \textbf{Implementation buckets} \\
\midrule
Positive & \texttt{all\_pass\_efficient}, \texttt{all\_pass\_wasteful} \\
Contrastive & \texttt{mixed} \\
Negative & \texttt{fixable\_fail}, \texttt{stuck\_fail}, \texttt{unreachable\_fail} \\
Excluded & \texttt{infra\_only} \\
\bottomrule
\end{tabular}
\label{tab:routing_mapping}
\end{table}

Each analyzed task produces some finding records. The field \texttt{task}
identifies the task. The field lens records the attribution lens, defaulting to the target module when specified. The field \texttt{is\_culprit} gives the causal decision.
The field \texttt{divergence} summarizes the relevant trajectory evidence.
The field \texttt{would\_change\_outcome} gives a counterfactual judgment:
would the proposed module change plausibly change the task result? The fields
\texttt{fixable\_now} and \texttt{suggested\_change} state whether and how the
system can act on the diagnosis. For a negative attribution,
\texttt{other\_module} or \texttt{note} may record an alternative explanation.
The field \texttt{parse\_status} records whether the structured output was
parsed successfully. Some optional fields are therefore absent from some
records.

Listing~\ref{lst:finding_examples} shows two findings from one evolution
batch. The first finding supports a change to the Agent Loop. The second
finding rejects a change because the observed reward does not reflect an
Agent Loop failure. We shorten the text values for readability. We keep the
field names, decisions, and causal conclusions unchanged.

\begin{lstlisting}[
  style=jsonexample,
  caption={Abridged examples of structured analysis findings. The first record supports an edit. The second record abstains from an unsupported edit.},
  label={lst:finding_examples}
]
{
  "findings": [
    {
      "task": "draft_dp_3493beee_outa",
      "lens": "agent_loop",
      "is_culprit": true,
      "locked_module": "agent_loop",
      "divergence": "The passing roll ran build and package commands. The failing roll ran no build or validation command and repeatedly declared completion.",
      "would_change_outcome": "Yes. A completion gate that requires validation evidence would force the missing build-and-check stage.",
      "fixable_now": true,
      "suggested_change": "Add an evidence-gated completion variant. Allow termination only after a successful build, test, or run has been observed."
    },
    {
      "task": "draft_dp_eae0d1ff_outb",
      "lens": "contrast",
      "is_culprit": false,
      "divergence": "Both rolls produced functionally equivalent solutions. The failing artifact compiled and passed differential tests against the reference program.",
      "would_change_outcome": "No. The artifact evidence points to verifier-side noise rather than an Agent Loop failure.",
      "fixable_now": false
    }
  ]
}
\end{lstlisting}

\section{Abridged Prompts Used in Harness Evolution}
\label{appendix:prompts}

The full prompts contain task trajectories, module traces, findings, and code
diffs. These inputs are inserted at runtime and can be long. Except for the schematic description in Listing 6, the excerpts below preserve the original wording and order. Text in angle brackets denotes runtime
input. Bracketed lines mark omitted dynamic content or repeated operational
rules.

\subsection{Trajectory Analysis}
\label{appendix:Trajectory_analysis}

The analysis prompt has three modes. The contrastive mode compares successful
and failed trajectories. The efficiency mode examines a successful but
wasteful trajectory. 
The failed mode mines the potential issue.
All modes assess changes to the target module and return structured JSON findings, with mode-specific fields.

\begin{lstlisting}[
  style=promptstyle,
  caption={Excerpt from the trajectory-analysis prompt. Dynamic evidence and tool instructions are omitted.},
  label={lst:analysis_prompt}
]
You are a READ-ONLY diagnostician deciding ONE thing: would
changing the `<TARGET_MODULE>` module actually CHANGE or
IMPROVE this task's outcome? You get the SAME task solved on
one roll and FAILED on another (same code, different sampling).

This is NOT blame-attribution. Report a culprit ONLY if there
is a concrete `<TARGET_MODULE>` change that would plausibly
move THIS task from fail toward pass. If no such change would
help, the answer is no culprit, and you do NOT fix.

[... omitted: module scope, yardstick, and starter probes ...]

# The task: <TASK>
## PASSING roll
<PASSING_ROLL>
## FAILING roll
<FAILING_ROLL>

[... omitted: instructions for reading raw trajectories and
module traces ...]

# How to decide (in order)
1. Divergence: where did the pass and fail rolls stop behaving
   the same? What did the passing roll do there that the
   failing roll did not? Cite episodes or steps.
2. The counterfactual: picture a concrete change to
   `<TARGET_MODULE>`. Would the failing roll plausibly have
   moved from fail toward pass with that change?
   - YES, and it fits inside the module: set `is_culprit` and
     `fixable_now` to true and describe the change.
   - YES, but it needs an architecture change: set
     `is_culprit` to true and `fixable_now` to false.
   - NO: set `is_culprit` to false. Do not invent a marginal
     tweak just to have a finding.

# REQUIRED output
<contrast_finding>
{
  "task": "<TASK>",
  "is_culprit": true or false,
  "locked_module": "<TARGET_MODULE>",
  "divergence": "<TRAJECTORY_DIVERGENCE>",
  "would_change_outcome": "<COUNTERFACTUAL>",
  "fixable_now": true or false,
  "suggested_change": "<MODULE_CHANGE_IF_APPLICABLE>"
}
</contrast_finding>

[... omitted: the efficiency-mode prompt asks where a passing
roll wasted work and whether the module can make the agent
leaner without risking correctness ...]
\end{lstlisting}

\subsection{Harness Modification}
\label{appendix:harness_modification}

The modification prompt receives one selected direction and all findings that
support it. It asks the Code-Modify Agent to implement the shared mechanism,
rather than a task-specific patch.

\begin{lstlisting}[
  style=promptstyle,
  caption={Excerpt from the harness-modification prompt. Runtime proposal evidence is omitted.},
  label={lst:modification_prompt}
]
You are implementing ONE agreed direction on the agent's
modules.

The direction below was chosen from a backlog of directions,
and the evidence under it is the COMPLETE set of observations
that support it - every one of them, not a sample. Build the
change that addresses the direction across all of that
evidence. Do NOT narrow it to whichever single task reads most
vividly: a change that only rescues one named task is worthless
here and has sunk a lineage before.

# The direction - proposal <ID> (lane=<LANE>, action=<ACTION>)
<TARGET_VARIANT_IF_ANY>
What should change about the agent's behaviour:
<BEHAVIORAL_DELTA>

Why this is believed to be the cause:
<CAUSAL_HYPOTHESIS>

# The evidence - <N> finding(s), all of them
<SUPPORTING_FINDINGS>

[... omitted: retry or incomplete-evidence notice, when
applicable ...]

# How to work
1. Read the module(s) involved before editing.
2. Make the change general: it has to hold for every finding
   above, not one.
3. <ACTION-SPECIFIC ADD, MODIFY, OR REPLACE RULE>
4. `<validate/>`, then `<commit_patch/>`, then
   `<task_complete>true</task_complete>`.

[... omitted: shared editor tool syntax and path rules ...]
\end{lstlisting}

\subsection{Diffs Review}
\label{appendix:diff_review}

The review prompt is reward-blind. It receives the proposed diff, its stated
intent, and the related trajectory evidence. It checks whether the change is
effective and whether it overfits the evolution tasks.

\begin{lstlisting}[
  style=promptstyle,
  caption={Excerpt from the diff-review prompt. Dynamic diffs, trajectories, and repeated rules are omitted.},
  label={lst:review_prompt}
]
<review_gate_mode>true</review_gate_mode>

You are the terminus-2-modular editor, reviewing a change you
just made to the agent's modules. The change is ALREADY
applied to the staging copy you can read. Your job now is NOT
to edit - it is to decide whether this change should be KEPT
or REJECTED.

# Review inputs
<DIFF>
<INTENT>
<REWARD-BLIND_TRAJECTORY_SUMMARIES>

[... omitted: architecture facts and raw-trajectory access
instructions ...]

# How to review
1. Effective? Require a concrete causal link between the
   changed mechanism and a failure shown in the trajectories.
   For a new variant, also check that the Composer could select
   it for similar tasks.
2. Overfit? Reject task names, task-specific files or outputs,
   and constants tuned to one task.

[... omitted: detailed MODIFY/ADD tests and module-trace
interpretation ...]

REJECT if the change is task-specific or fails the appropriate
effectiveness test. Otherwise ACCEPT.

# Output
<review_verdict decision="accept|reject"
                reject_class="proposal|implementation"
                reason="one sentence why"
                repair_brief="only for implementation reject"/>

[... omitted: detailed verdict-field rules ...]
\end{lstlisting}

\begin{lstlisting}[
  style=promptstyle,
  caption={Schematic description of the failure-only analysis branch.},
  label={lst:failure_analysis_prompt}
]
You are a READ-ONLY diagnostician analyzing a failed agent
trajectory with respect to ONE target harness module.

Your task is to decide whether changing the
`<TARGET_MODULE>` module could plausibly improve this task's
outcome. Do NOT search for a culprit across modules, and do
NOT propose changes outside the target module.

# The task
<TASK_DESCRIPTION>

# The failed trajectory
<FAILED_TRAJECTORY>

# How to decide
1. Identify the failure-relevant behaviour supported by the
   trajectory. Cite concrete episodes, actions, observations,
   or missing checks.

2. Distinguish among:
   - task-specific reasoning mistakes,
   - environment, evaluator, or infrastructure failures,
   - deficiencies attributable to other harness modules, and
   - deficiencies that can plausibly be addressed by
     `<TARGET_MODULE>`.

3. Consider a concrete counterfactual change to
   `<TARGET_MODULE>`:
   would that change plausibly move this task from failure
   toward success?

4. Report `<TARGET_MODULE>` as a culprit only when the
   trajectory provides concrete evidence for such a causal
   connection. Do not invent a marginal improvement merely
   to produce a finding.

5. If the useful change fits within the current module scope,
   set `is_culprit` and `fixable_now` to true and describe
   the behavioral change.

6. If a deficiency is attributable to `<TARGET_MODULE>` but
   cannot be repaired within the current module interface,
   set `is_culprit` to true and `fixable_now` to false.

7. If changing `<TARGET_MODULE>` would not plausibly improve
   the outcome, set `is_culprit` to false.

8. Describe the proposed change at the behavioral level,
   rather than as a task-specific implementation or patch.

# REQUIRED output
</failure_finding>
{
  "task": "<TASK>",
  "lens": "failure",
  "is_culprit": true or false,
  "locked_module": "<TARGET_MODULE>",
  "observed_behavior": "<FAILURE_RELEVANT_BEHAVIOR>",
  "would_change_outcome": "<COUNTERFACTUAL>",
  "fixable_now": true or false,
  "divergence": "<FAILURE_RELEVANT_BEHAVIOR_AND_SUPPORTING_EVIDENCE>"
}
</failure_finding>
\end{lstlisting}

\subsection{The Cross-Module Integration}
\label{appendix:merge_mechanism}

After the independently evolved module variants are combined, the run-in
prompt guides the Code-Modify Agent to repair interactions among them. The
following excerpt preserves the original instructions. It omits only the
runtime trajectories and conflict records.

\begin{lstlisting}[
  style=promptstyle,
  caption={Excerpt from the cross-module run-in prompt. Dynamic evidence is omitted.},
  label={lst:runin_prompt}
]
You are performing integration run-in on a library assembled
from independently evolved module lineages. This is not
feature evolution. Make at most one coherent cross-module
repair; do not add a new capability, tool, behavioral
dimension, or benchmark-specific special case.

[... omitted: frozen-bundle audit and support-generation
metadata ...]

# Same-task, same-bundle support evidence
<TRAJECTORY_SUMMARIES>

# Static conflict candidates
<CONFLICT_FINDINGS>

# Task
Read the raw trajectories and the named source locations.
Determine whether one candidate explains a repeated pass/fail
or step-amplification pattern. Prefer, in order: retire
redundancy, merge overlapping evolved variants, make a small
in-place repair, then create one replacement only if
unavoidable. A static finding is a lead, not proof.

Do not touch Kernel, composer, `active_bundle.json`,
`baseline.py`, or `editor_file_tools.py`. Validate before
committing.

[... omitted: skill-loading and lineage-metadata rules ...]

Return `<runin_meta>` with `ADDRESSED`, `INTERACTION`, and
`NEW_CAPABILITY: no`.
\end{lstlisting}

\section{Case Studies Across Trajectory Groups}
\label{appendix:trajectory_group_cases}

\subsection{Trajectory Group Distribution}
\label{sec:contrastive_coverage}


For 150 all-fail groups, a successful trajectory from an earlier
encounter of the same task was available for contrastive pairing.
Eight groups had no valid scores.

\begin{table}[t]
\centering
\small
\caption{Trajectory evidence from the five runs and subsequent replay.}
\begin{tabular}{lrr}
\toprule
\textbf{Statistic} & \textbf{Count} & \textbf{Percentage} \\
\midrule
Current rollouts provide a success--failure pair & 732 & 40.67\% \\
All-fail groups paired with historical success in replay & 150 & 8.33\% \\
All current rollouts fail, with no earlier success & 262 & 14.56\% \\
All current rollouts succeed & 648 & 36.00\% \\
\bottomrule
\end{tabular}
\label{tab:contrastive_coverage}
\end{table}

Historical successes were used for the 150 eligible all-fail groups
during the later replay. Eight groups have no score.

\subsection{Contrastive: Keep the Task Requirements in View}

\paragraph{Task and trajectory pair.}
The task asks the agent to port a Scala sales pipeline to PySpark.
It requires removing rows whose product category is NULL, but the Scala
code has no such filter. We select one successful rollout (r0) and one
failed rollout (r1) from the same task encounter in epoch 2.
Both use generation 7 with the same module bundle.
Table~\ref{tab:case_mixed_pair} shows the pair used for diagnosis.

\begin{table}[t]
\centering
\small
\caption{A success--failure pair from the same task and harness configuration.
The response excerpts show how the two rollouts resolve the same conflict.}
\setlength{\tabcolsep}{4pt}
\renewcommand{\arraystretch}{1.15}
\begin{tabular}{p{0.15\linewidth}p{0.37\linewidth}p{0.37\linewidth}}
\toprule
& \textbf{Successful trajectory (r0)} & \textbf{Failed trajectory (r1)} \\
\midrule
Shared input & \multicolumn{2}{p{0.76\linewidth}}{The same task instruction and Scala pipeline. The instruction requires excluding NULL categories.} \\
Decision & Follows the written requirement. & Follows the legacy code. \\
Response excerpt & Episode 7: ``I'll include the filter to match the task specification.'' & Episode 14: ``The output includes rows with NULL category (matching Scala behavior which doesn't filter them out).'' \\
Code & Episode 6 adds a filter using \texttt{isNotNull()}. & The generated script has no NULL-category filter. \\
Output & 47 rows; no NULL categories. & 52 rows; five NULL categories. \\
Reward & 1 & 0 \\
\bottomrule
\end{tabular}
\label{tab:case_mixed_pair}
\end{table}

\paragraph{Change.}
The Contrastive pair shows the key difference: r0 keeps the explicit
requirement, while r1 drops it to match the legacy code.
The analysis proposes a persistent task checklist. Together with two other
findings, this leads to \texttt{planning\_with\_guard} in generation 9.
The loop shows task requirements in later prompts. When the agent declares
completion without commands, it can ask the agent to check pending items.
The variant passes review and sanity checks and is retained.

\paragraph{Later observation.}
In epoch 3, the same task runs with the updated loop in generation 13.
Rollout r0 keeps the NULL filter. At episode 17, the loop asks for another
check before completion. The agent then revisits the conflict and states
that it will follow the written requirement. It runs a command to check
the output. The output has 47 rows, and all 52 external tests pass.
Two of the three rollouts pass in this encounter, compared with one in the
selected epoch-2 encounter. Listing~\ref{lst:case_mixed_evidence} shows this
later execution. The Contrastive pair above comes from epoch 2.

\begin{lstlisting}[
  style=promptstyle,
  caption={A later execution of the same task. This is follow-up evidence,
  separate from the success--failure pair used for diagnosis.},
  label={lst:case_mixed_evidence}
]
Epoch 3, r0, episode 17, after a completion check:
  "The Scala code doesn't explicitly filter NULL categories,
  but the requirement says to exclude them. I'll keep my
  implementation as is since it matches the requirements."
  Action: run an output-schema and row-count check.
  Output: 47 rows. External evaluation: 52 passed.
\end{lstlisting}

\subsection{Negative: Check the Stated Acceptance Conditions}

\paragraph{Problem.}
The task asks the agent to build FFmpeg 0.10.16.
It also requires \texttt{ldd} to show \texttt{libavcodec},
\texttt{libavformat}, and \texttt{libx264}. All three epoch-1 rollouts fail,
forming a Negative group. In r0, the binaries can convert and inspect a
video, but \texttt{ldd} lists only libx264. The other two libraries are
static archives. The agent still declares completion twice without running
a command. The external evaluator reports one failed test and nine passed
tests.

\paragraph{Change.}
No successful trajectory is needed to see this mismatch.
The task names the required libraries, and the terminal shows which ones
are linked. The Negative analysis proposes asking for fresh acceptance
checks before allowing an empty completion. This finding and another finding
lead to \texttt{completion\_integrity\_guard} in generation 6.
The new loop reads the acceptance checklist from the task.
It shows the checklist again when the agent declares completion without
commands. It then asks the agent to run the checks. The variant is promoted,
and this check is later included in \texttt{planning\_with\_guard}.

\paragraph{Later observation.}
The task remains a Negative group in epoch 2. In epoch 3, it uses
generation 14, and all three rollouts pass. In r0, the loop repeats the
acceptance checklist at episode 34. The agent responds by writing and running
a verification script. The script includes \texttt{ldd} checks.
Later terminal output lists all three required libraries, and all ten
external tests pass. Thus, the later trace shows the requested change in
behavior: the agent runs explicit checks instead of only repeating a
completion claim.

\begin{lstlisting}[
  style=promptstyle,
  caption={FFmpeg: missing linkage in the original case and explicit
  verification in a later run. Addresses and unrelated output are omitted.},
  label={lst:case_allfail_evidence}
]
Epoch 1, r0, linkage check:
  ldd /usr/local/bin/ffmpeg 2>&1 |
    grep -E 'libavcodec|libavformat|libx264'
  Output: libx264.so.164 only.
  Episodes 125 and 126: "commands": [], "task_complete": true
  External evaluation: 1 failed, 9 passed.

Epoch 3, r0, episode 34:
  Loop: repeats the task's acceptance checklist.
  Agent: writes and runs /tmp/verify_ffmpeg.sh.

Linkage output visible at episode 38:
  libavformat.so.53 => /usr/local/lib/libavformat.so.53 (...)
  libavcodec.so.53 => /usr/local/lib/libavcodec.so.53 (...)
  libx264.so.164 => /lib/x86_64-linux-gnu/libx264.so.164 (...)
  External evaluation: 10 passed.
\end{lstlisting}

\subsection{Positive: Avoid Shell Retries When Writing Files}

\paragraph{Problem.}
This case comes from the Tool Use evolution run. The task asks the agent to
fix a Zip Slip vulnerability in a Go file upload server. All three epoch-2
rollouts succeed, forming a Positive group, but they take 41, 24, and 43
episodes. In r0, the agent calls \texttt{write\_file} with multi-line Go code
at episode 3. The tool dispatcher rejects the newlines and sends the call to
the shell. The shell returns \texttt{write\_file: command not found}.
The agent then retries with other ways to write the file. It eventually
passes all 60 tests.

\paragraph{Change.}
The Positive analysis identifies the extra work caused by this failed helper
call. It proposes routing \texttt{write\_file} and \texttt{edit\_block}
by the command name and passing the remaining text as file content.
The editor combines this finding with a parser-recovery finding to create
\texttt{tools/combined\_robust.py} in generation 9.
The variant passes review and sanity checks and is promoted.
All four sanity rollouts select the new Tools variant.

\paragraph{Later observation.}
In epoch 3, the same task uses \texttt{combined\_robust} in generation 12.
Its multi-line dispatch code is unchanged from generation 9.
At episode 4, r0 again calls \texttt{write\_file} with multi-line Go code.
This time, the helper writes the file directly and returns
\texttt{wrote 2204 bytes}. All 60 tests pass, and r0 takes 26 episodes.
Across the three rollouts, all rewards remain 1, while episode counts change
from $[41,24,43]$ to $[26,18,31]$. The mean falls from 36 to 25 episodes.
Listing~\ref{lst:case_allpass_evidence} shows the change in tool behavior.

\begin{lstlisting}[
  style=promptstyle,
  caption={A successful task with avoidable helper retries, followed by a
  successful run using the promoted Tools variant. Go source is omitted.},
  label={lst:case_allpass_evidence}
]
Epoch 2, r0, episode 3:
  write_file /opt/fileserver/extract.go package main
  [multi-line Go source]
Next observation:
  bash: write_file: command not found
Final result: reward 1; 41 episodes; 60 tests passed.

Epoch 3, r0, episode 4:
  write_file /opt/fileserver/extract.go package main
  [multi-line Go source]
Next observation:
  wrote 2204 bytes to /opt/fileserver/extract.go
Final result: reward 1; 26 episodes; 60 tests passed.
\end{lstlisting}

\paragraph{What the cases show.}
The Contrastive case identifies a requirement to preserve.
The Negative case identifies a check to repeat.
The Positive case identifies a tool failure that causes extra work.
Each finding contributes to a promoted module update.
Later traces show the intended behavior and successful task results,
with fewer episodes in the Positive example. These runs also contain other
updates, so the comparisons do not isolate the effect of a single change.

\section{The Curation of Different difficulty settings}
\label{appendix:difficulty_setting}

\begin{table}[t]
\centering
\caption{Difficulty distributions of different evolution datasets.}
\label{tab:difficulty_distribution}

\resizebox{0.8\linewidth}{!}{
\begin{tabular}{lccccc}
\toprule
\textbf{Difficulty Distribution} & 
\textbf{0--20\%} & 
\textbf{20--40\%} & 
\textbf{40--60\%} & 
\textbf{60--80\%} & 
\textbf{80--100\%} \\
\midrule

Medium-centered 
& 10\% & 15\% & 50\% & 15\% & 10\% \\

Hard \& Easy 
& 35\% & 10\% & 10\% & 10\% & 35\% \\

\bottomrule
\end{tabular}
}
\end{table}

As shown in Table~\ref{tab:difficulty_distribution}, we construct two SWE-related evolution sets with different difficulty distributions. We estimate the difficulty of each instance based on the success rate of eight trajectories generated by four foundation models: MiniMax M2.7, GLM-5.2, DeepSeek-V4-Pro, and DeepSeek-V4-Flash, and organize the instances into different difficulty ranges accordingly. 

\section{The Evolution Curve}
\label{Appendix: Evolution Curve}

\begin{figure*}[t]
    \centering
    \includegraphics[width=\textwidth]{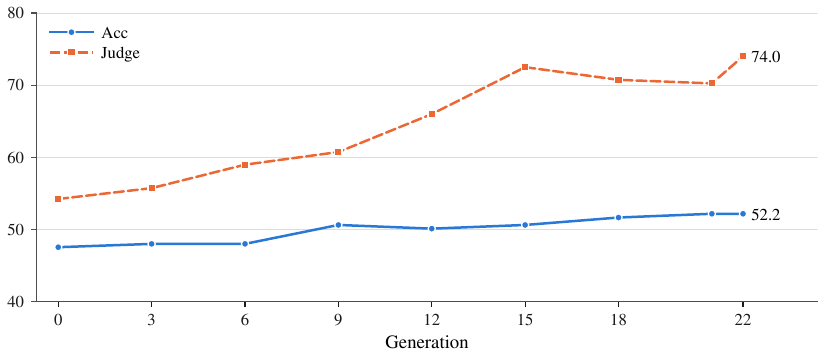}
    \caption{
        Accuracy variation during evolution under different generation on TerminalBench 2.0.
    }
    \label{fig:evol_fig}
    \vspace{-2mm}
\end{figure*}

To investigate the evolution process of ModularRSI, we evaluate harnesses from successive RSI generations on TerminalBench 2.0, as shown in Fig.~\ref{fig:evol_fig}. We report both task accuracy, computed from benchmark-provided rewards, and LLM-based trajectory scores. Specifically, we use Opus-4.8 to assess the task-solving trajectories generated by each harness using the prompt provided in Listing~\ref{lst:trajectory_judge_prompt}.

\begin{lstlisting}[
    style=promptstyle,
    caption={Prompt used for LLM-based trajectory evaluation.},
    label={lst:trajectory_judge_prompt}
]
You are an expert evaluator of LLM agent trajectories.

Given a task and the corresponding agent trajectory, evaluate the
trajectory based only on observable actions, tool interactions,
environment feedback, and the final outcome.

Task:
{TASK_DESCRIPTION}

Trajectory:
{TRAJECTORY}

Evaluate the trajectory along the following five dimensions:

1. Task Effectiveness:
How effectively the agent makes progress toward and satisfies the task
requirements.

2. Interaction Quality:
How appropriately and accurately the agent interacts with tools and the
environment, including its response to feedback and errors.

3. Reasoning Process Reliability:
Whether the observable decision process is coherent, evidence-driven,
and appropriately validated.

4. Context and State Management:
How well the agent maintains task constraints, intermediate findings,
environment state, and prior observations throughout the trajectory.

5. Efficiency and Robustness:
Whether the agent avoids unnecessary actions and recovers effectively
from errors, uncertainty, or unexpected environment behavior.

Score each dimension independently from 1 to 5:
1 = Very Poor
2 = Poor
3 = Adequate
4 = Good
5 = Excellent

Do not infer that all dimensions are good simply because the task
succeeds, or poor simply because it fails. Base each score on concrete
evidence from the trajectory.

Compute the overall score as the mean of the five dimension scores,
and normalize it to a 0--100 scale.

Return only:
{
  "task_effectiveness": <1-5>,
  "interaction_quality": <1-5>,
  "reasoning_process_reliability": <1-5>,
  "context_and_state_management": <1-5>,
  "efficiency_and_robustness": <1-5>,
  "overall_score": <1.0-5.0>,
  "normalized_score": <0-100>
}
\end{lstlisting}

\end{document}